\documentclass{article}

\usepackage[preprint,nonatbib]{neurips_2022}

\usepackage[utf8]{inputenc} 
\usepackage[T1]{fontenc}    
\usepackage{hyperref}       
\usepackage{url}            
\usepackage{booktabs}       
\usepackage{amsfonts}       
\usepackage{nicefrac}       
\usepackage{microtype}      
\usepackage{xcolor}         
\usepackage{graphicx}
\usepackage{subfigure}
\usepackage{float}
\usepackage{multirow}
\usepackage{wrapfig}
\usepackage{caption}
\usepackage[ruled,vlined]{algorithm2e}
\usepackage{amsmath}
\usepackage{algorithm, algorithmic}
\usepackage{authblk}

\title{Mask-guided Vision Transformer (MG-ViT) for Few-Shot Learning}

\author[1]{Yuzhong Chen}
\author[1]{Zhenxiang Xiao}
\author[3]{Lin Zhao}
\author[4]{Lu Zhang}
\author[3]{Haixing Dai}
\author[3]{David Weizhong Liu}
\author[3]{Zihao Wu}
\author[2]{Changhe Li}
\author[2]{Tuo Zhang}
\author[3]{Changying Li}
\author[4]{Dajiang Zhu}
\author[3]{Tianming Liu}
\author[1]{Jiang Xi \thanks{Corresponding author: xijiang@uestc.edu.cn}}
\affil[1]{University of Electronic Science and Technology of China}
\affil[2]{Northwestern Polytechnical University}
\affil[3]{University of Georgia}
\affil[4]{University of Texas at Arlington}

\begin{document}

\maketitle

\begin{abstract}
Learning with little data is challenging but often inevitable in various application scenarios where the labeled data is limited and costly. Recently, few-shot learning (FSL) gained increasing attention because of its generalizability of prior knowledge to new tasks that contain only a few samples. However, for data-intensive models such as vision transformer (ViT), current fine-tuning based FSL approaches are inefficient in knowledge generalization and thus degenerate the downstream task performances. In this paper, we propose a novel mask-guided vision transformer (MG-ViT) to achieve an effective and efficient FSL on ViT model. The key idea is to apply a mask on image patches to screen out the task-irrelevant ones and to guide the ViT to focus on task-relevant and discriminative patches during FSL. Particularly, MG-ViT only introduces an additional mask operation and a residual connection, enabling the inheritance of parameters from pre-trained ViT without any other cost. To optimally select representative few-shot samples, we also include an active learning based sample selection method to further improve the generalizability of MG-ViT based FSL. We evaluate the proposed MG-ViT on both Agri-ImageNet classification task and ACFR apple detection task with gradient-weighted class activation mapping (Grad-CAM) as the mask. The experimental results show that the MG-ViT model significantly improves the performance when compared with general fine-tuning based ViT models, providing novel insights and a concrete approach towards generalizing data-intensive and large-scale deep learning models for FSL.
\end{abstract}

\section{Introduction}\label{submission}
Deep neural networks (DNNs) have achieved great success in many computer vision tasks with a large amount of labeled data. However, learning with little data is often inevitable in a variety of application scenarios in which the labeled data is limited and costly. Recently, few-shot learning (FSL) has attained increasing interest to reconcile the demand and scarcity of large-scale labeled data in DNNs because of its generalizability of prior knowledge to new tasks given only a few samples. Existing literature studies have devoted extensive efforts to improving FSL from three aspects \cite{wang2020generalizing}: data augmentation \cite{yang2021free,yan2022budget}, model design \cite{ma2021few}, and algorithm development \cite{zhang2021rethinking}, and obtained promising results. By simply freezing the backbone and merely fine-tuning the last few layers, previous fine-tuning based FSL approaches have achieved promising performances \cite{nakamura2019revisiting,wang2020frustratingly,cai2020cross}.

However, for data-intensive models with even more parameters, e.g., Vision Transformer (ViT) \cite{dosovitskiy2020image}, current fine-tuning based FSL approaches are inefficient in knowledge generalization. For example, in FSL scenario, ViT has better performance than ResNet only when being pretrained on large dataset\cite{dosovitskiy2020image}. This is probably because the data-intensive models, such as ViT, do not inherently encode the inductive biases that are useful for smaller datasets and thus require a large amount of labeled data to figure out the underlying modality-specific rules \cite{dosovitskiy2020image, khan2021transformers}, resulting in unsatisfying performance when the labeled data is limited. Therefore, figuring out how to efficiently generalize the domain knowledge of data-intensive models for FSL is still an open question and worth more effort to explore.


\begin{wrapfigure}{r}{6cm}
\includegraphics[width=6cm]{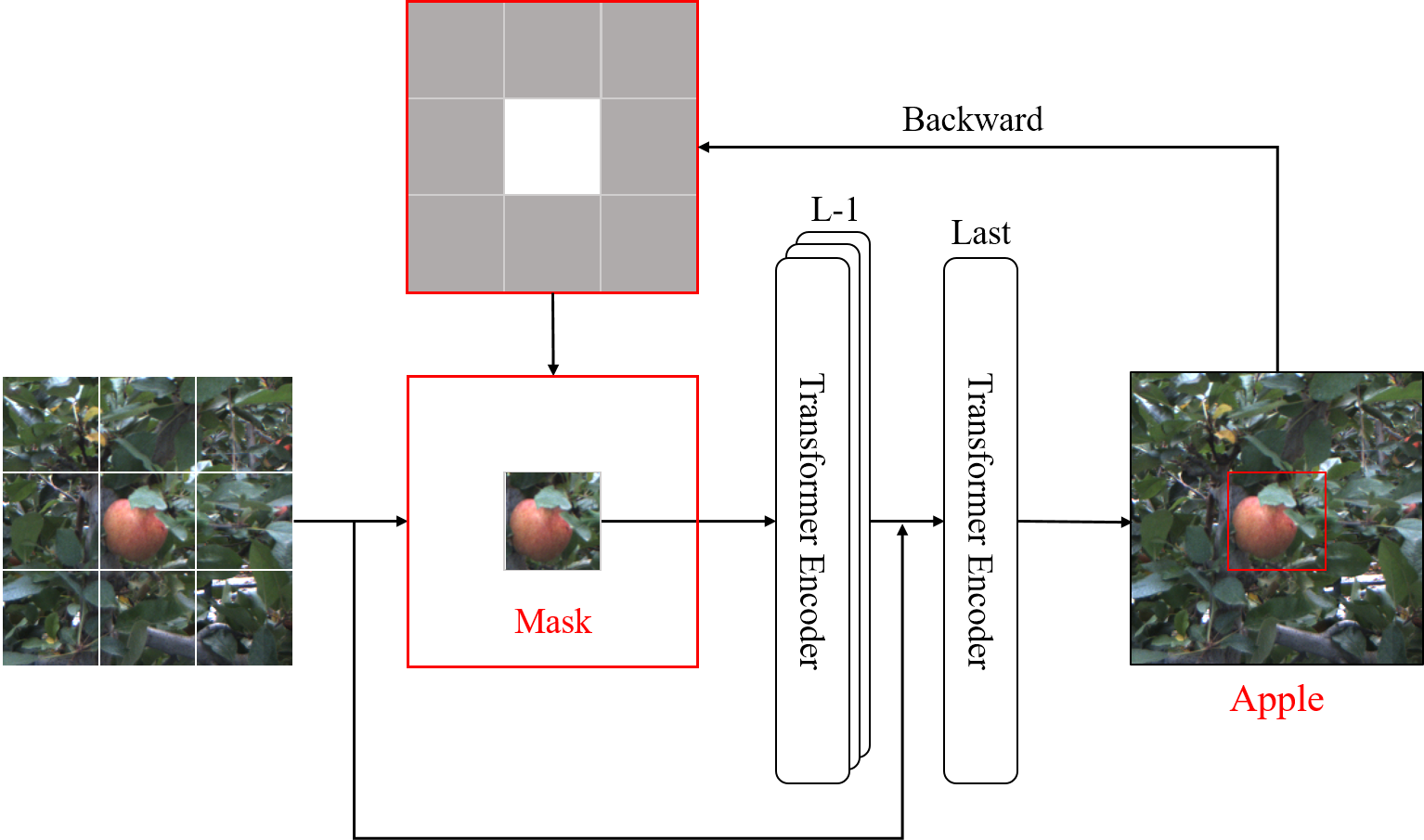}
\captionof{figure}{Mask operation to select most task-relevant and discriminative image patches for MG-ViT model.}\label{figure_1}
\end{wrapfigure}
In this paper, we propose a novel mask-guided vision transformer (MG-ViT) to effectively adapt the domain knowledge of pre-trained ViT to FSL tasks. The key idea of MG-ViT is to apply a mask on input image patches before the first transformer encoder layer to screen out the task-irrelevant patches and to guide ViT to focus on task-relevant and discriminative ones during FSL (\ref{figure_1}) as background information is harmful for few-shot learning \cite{luo2021rectifying}. The most task-relevant and discriminative patches are identified from the salience map by calculating the gradient-weighted class activation mapping (Grad-CAM) \cite{selvaraju2017grad} of the most similar images in the source dataset with the target dataset. The proposed mask operation can take better advantage of the prior knowledge from the source domain and reduce the deviation between the source and target domains. Moreover, since the visible patches that are not masked retain task-relevant information but lose global information, especially the position information in the image, we add a residual connection between the first and last encoder layers of ViT to retain the global information. MG-ViT follows a two-stage training scheme for FSL \cite{wang2020frustratingly}. The first stage is to train on a base dataset only with the vanilla ViT, and the second stage is to fine-tune the model on both novel and base datasets with the proposed MG-ViT. Particularly, MG-ViT only introduces the above-mentioned mask operation and residual connection to ViT, and is able to inherit parameters from pre-trained ViT without any other costly pre-training.

To further improve the generalizability of MG-ViT based FSL, we additionally include an active learning based sample selection method \cite{cohn1994improving,chitta2021training}. The rationale is that since general FSL aims to predict a large number of unlabeled samples by learning from a limited number of labeled data \cite{wang2020generalizing}, active learning may improve the model learning by labeling representative samples and further improve the model prediction ability on other unlabeled samples \cite{cohn1994improving,konyushkova2017learning}. Therefore, we introduce a cluster-based sample selection method to optimally select representative few-shot samples.

We evaluate the proposed MG-ViT on both Agri-ImageNet image classification task and ACFR apple detection task, and conduct extensive comparisons with the general fine-tuning based ViT models. The experimental results demonstrate that the proposed MG-ViT model significantly improves the task performance when compared with general fine-tuning based ViT models.

In general, the main novelties and contributions of our work are:

\begin{itemize}
\item We introduce an elegant mask operation on image patches to guide the ViT to focus on task-relevant and discriminative patches during FSL, add a residual connection to retain the global features of visible image patches, and maintain the other structures of ViT, thus allowing improvement of model effectiveness with inheritance of parameters from pre-trained ViT without any other cost.
\item We include an active learning based sample selection method to optimally select representative few-shot samples, which allows further improvement of FSL on the proposed MG-ViT.
\item We propose an effective framework of FSL on data-intensive models (ViT in this study) and achieve excellent performances on both image classification and object detection tasks, thus providing novel insights in generalizing data-intensive and large-scale deep learning models for FSL.
\end{itemize}

\section{Related work}

\paragraph{Vision Transformer}
ViT \cite{dosovitskiy2020image} is transferred from the structure of transformer in natural language processing (NLP) tasks \cite{vaswani2017attention}. Recently, several refined ViT models such as DeiT \cite{touvron2021training}, CeiT \cite{yuan2021incorporating}, local ViT \cite{li2021localvit} and NesT \cite{zhang2021aggregating} are introduced with useful strategies including knowledge distillation, depth-wise convolution and tree-like structure, and achieve improved model performance in the vision task.

However, the data-intensive ViT is difficult to quickly adapt to the target domain with a small amount of labeled data. By using a distillation approach \cite{chitta2021training}, smoothing the loss landscapes at convergence \cite{chen2021vision}, or incorporating CNNs like CCT \cite{hassani2021escaping} and NesT \cite{zhang2021aggregating}, ViT can reduce the demand for large sample data to some extent. Moreover, MAE \cite{he2021masked} is proposed to mask most of the image patches and apply an unsupervised image reconstruction method to pre-train transformer to improve its generalizability for downstream tasks. However, these studies are still far from the requirement of fast adaption to the target domain for FSL.

\paragraph{Few-shot Learning}
Few-shot learning (FSL) is proposed for learning with only a few samples \cite{wang2020generalizing}. Meta-learning, which is also known as learning-to-learn, is a crucial approach for FSL \cite{vanschoren2018meta}. In recent studies for FSL, Zhang et al. \cite{zhang2021rethinking} proposed a novel absolute-relative learning paradigm to fully use the binary label and soft similarity information. Ma et al. \cite{ma2021few} designed an inverted pyramid network (IPN) with global and local stages to learn the support-query relation and precise query-to-class similarity embedding. Yang et al. \cite{yang2021free} calibrated the distribution of few sample class by transferring statistics from the classes with sufficient examples subject to a Gaussian distribution of each feature representation. For transformer based FSL, recent studies \cite{liu2020universal,gan2021transformer,chen2021sparse} merely fix the CNN-based feature extractor trained on base class and apply the attention mechanisms to exploit the correlation between query and support sets to perform classification on FSL, which does not take full advantage of the powerful learning representation of ViT. 

\paragraph{Active Learning}
Previous studies usually randomly select certain images from the dataset as training samples and have the model \emph{learn from examples} \cite{cohn1994improving}. However, by actively selecting a fixed number of training data, active learning is provably more powerful and with better generalizability \cite{cohn1994improving,konyushkova2017learning}. For example, Chitta et al. \cite{chitta2021training} designed an ensemble active learning method and achieved better performance on the test data with less training data. Therefore, active learning may improve FSL by labeling representative few-shot samples \cite{wang2020generalizing}. For example, Yan et al. \cite{yan2022budget} reported a better result on FSL using a Graph Convolution Network (GCN) based active learning data selection policy.

\section{Methods}
\begin{figure*}[t]
\begin{center}
\includegraphics[width=0.85\linewidth]{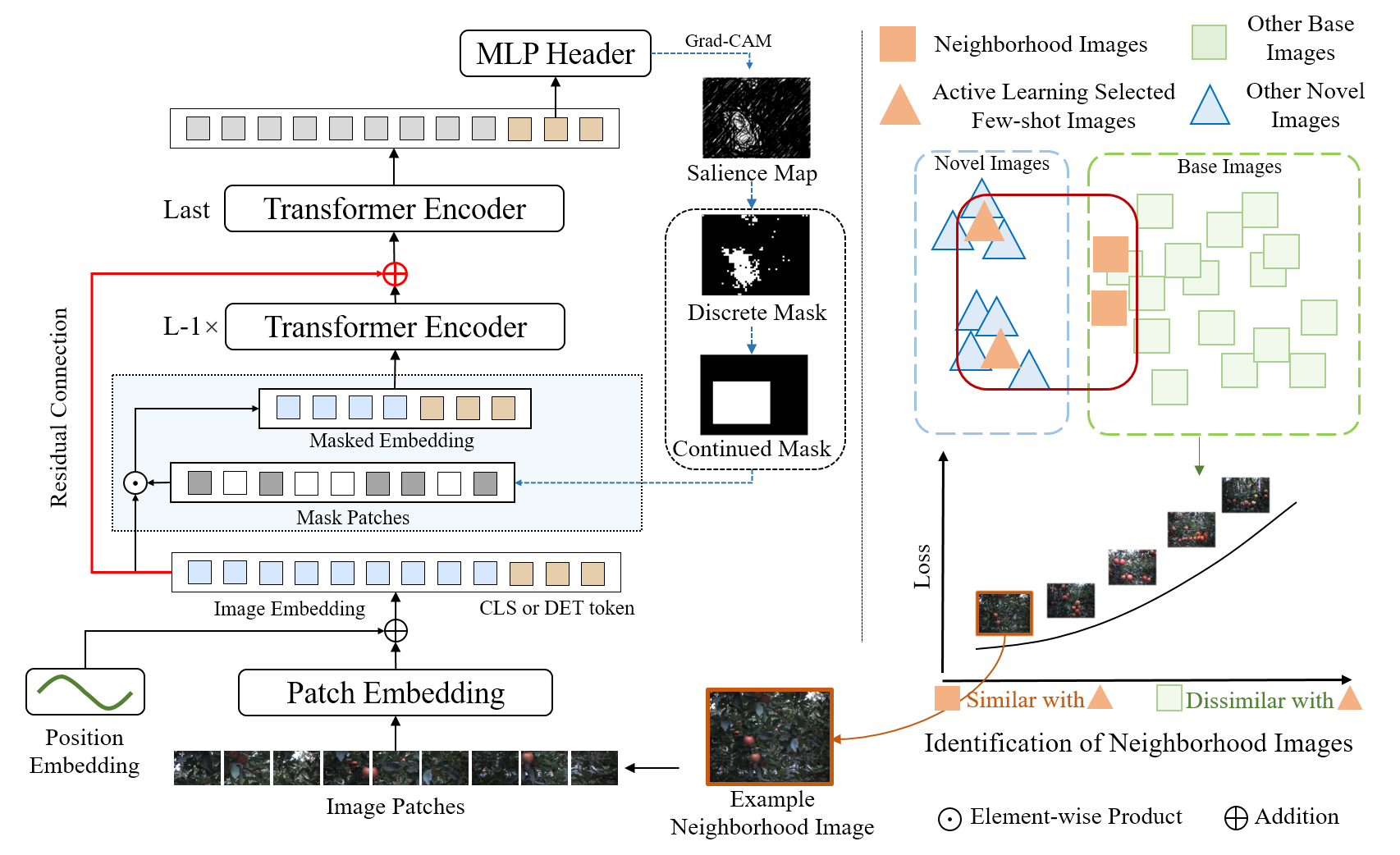}
\caption{The overall framework of MG-ViT for FSL. Left: Model structure of MG-ViT. The proposed additional mask operation is highlighted with \textcolor[RGB]{135,206,255}{blue} background and the residual connection is highlighted with \textcolor{red}{red} line. The continued or discrete image patch mask of task-irrelevant regions is generated based on the salience map by Grad-CAM on the patch embedding layer. Right: \textcolor[RGB]{244,158,98}{Orange triangles} are representative few-shot samples selected from the novel dataset by active learning. \textcolor[RGB]{244,158,98}{Orange squares} are identified neighborhood samples from the base dataset having lower loss values (i.e., higher similarity) with the representative few-shot samples. The identified neighborhood samples and representative few-shot samples highlighted in the red box are the training data during the joint fine-tuning stage. CLS: class token; DET: detect token.}
\label{figure_2}
\end{center}
\end{figure*}

The overall framework is illustrated in Section~\ref{figure_2}.  We first illustrate the definitions of our problem and ViT architecture in Section~\ref{background}. We next demonstrate the detailed structure of MG-ViT in Section~\ref{MG-ViT} and the generation of image patch mask in Section~\ref{mask}. We then introduce the identification of neighborhood image in Section~\ref{neighbor} and active learning based few-shot sample selection in Section~\ref{activelearning}. Lastly, we provide the overall training scheme in Section~\ref{training}.

\subsection{Problem Definition and Vision Transformer}
\label{background}

\paragraph{Problem Definition:} Given a base dataset $D_b=\{(x,y)\}$  where $y \in Y_b$ and a novel dataset $D_n =\{ (x,y)\} $ where $y \in Y_n$, $Y_b \cup Y_n =Y$, $ Y_b \cap Y_n = \emptyset$ where $Y$  denotes the whole set of class. The base dataset is with a large amount of labeled data while the novel dataset is with only a few labeled data. Our aim is to train a model with both base and novel datasets while achieving satisfying generalizability on the novel dataset for FSL. The common practice to evaluate the fast adaptation ability and generalizability of FSL model is to build a $N$-way-$K$-shot task, where $N$ is the number of classes and $K$ is the number of labeled data in the novel dataset. The model is trained with only $N\times K$ labeled data from the novel dataset.
The performance of FSL model is evaluated on the test split of the novel dataset. 

\paragraph{Vision Transformer:} ViT \cite{dosovitskiy2020image} receives an patch embedding sequence $x_{\scriptscriptstyle PATCH}\in \mathbb{R} ^{N\times D}$, where $\displaystyle N = \frac{H*W}{P^{2}}$ is the number of patches, $D$ is the output dimension, and $(H, W)$ and $(P, P)$ are the resolutions of the image and patch, respectively. For different downstream tasks, $x_{\scriptscriptstyle PATCH}$ concatenates different tokens, e.g., $x_{\scriptscriptstyle CLS}\in \mathbb{R} ^{1\times D}$ class token \cite{dosovitskiy2020image} for the image classification task and $x_{\scriptscriptstyle DET}\in \mathbb{R} ^{100\times D}$ detect token \cite{fang2021you} for the object detection task. To retain the position information of patches in the whole image, position embedding $\mathbf{P}  \in \mathbb{R} ^{(N+1+100)\times D}$ is also added to the concatenated inputs. Therefore, the input sequence $z_0$ of ViT with both class token and detect token is described as:
\begin{equation}
    z_0 = [x_{\scriptscriptstyle PATCH};x_{\scriptscriptstyle DET};x_{\scriptscriptstyle CLS}] + \mathbf{P}
\end{equation}
The encoder layer of Transformers consists of one multi-head self-attention (MSA) block and one multi-layer perceptron (MLP) block. LayerNorm (LN) and residual connections are applied before and after every block, respectively. Therefore, the output embedding of $l$-th layer is:
\begin{align}
    z_l' &= MSA(LN(z_{l-1}))+z_{l-1} ,\, l=1,\cdots,L \\
    z_l &= MLP(LN(z_l'))+z_l' ,\, l=1,\cdots,L
\end{align}
For different downstream tasks, the task-relevant tokens are fed into the task-special MLP header for final prediction. For image classification task:
\begin{equation}
    y_{class} = MLP(x_{\scriptscriptstyle CLS})
\end{equation}
For object detection task:
\begin{align}
    y_{class} \!&=\! [MLP_c(x_{\scriptscriptstyle DET}^1);\cdots;MLP_c(x_{\scriptscriptstyle DET}^{100})]\\
    y_{bbox} \!&=\! [MLP_b(x_{\scriptscriptstyle DET}^1);\cdots;MLP_b(x_{\scriptscriptstyle DET}^{100})]
\end{align}

\subsection{Mask-guided Vision Transformer}
\label{MG-ViT}
We introduce the detailed structure of proposed MG-ViT as follows. One of the core issues for FSL is to effectively adapt the prior knowledge learned from the source domain (base dataset) to the target domain (novel dataset). Inspired by He et al. \cite{he2021masked}, we add an image patch mask (detailed in Section~\ref{mask}) to the patch embeddings before the first encoder layer of transformer to screen out the task-irrelevant image patches and to guide ViT focusing on task-relevant and discriminative ones. Different from masking random patches in MAE \cite{he2021masked}, our design masks the task-irrelevant ones to focus on the task-relevant prior knowledge for better generalizability. Note that we only apply the mask operation on the base dataset but not on the novel dataset for two reasons. First, we want to make fully use of the few-shot sample information in the novel dataset for better FSL. Second, it is difficult to identify the important features within only a few samples which may lead to a noisy salience map. Therefore, the input of the first encoder layer $z_0^{masked}$ is:
\begin{equation}
    z_0^{masked} = [x_{\scriptscriptstyle PATCH}\odot Mask;x_{\scriptscriptstyle DET};x_{\scriptscriptstyle CLS}]
    +[P_{\scriptscriptstyle PATCH}\odot Mask;P_{\scriptscriptstyle DET};P_{\scriptscriptstyle CLS}]
\end{equation}
where $P_{\scriptscriptstyle PATCH}$, $P_{\scriptscriptstyle DET}$ and $P_{\scriptscriptstyle CLS}$ are the patch position embedding, detect position embedding and class position embedding in P, respectively. $\odot$ is the element-wise production.

Moreover, we introduce a residual connection between the first and last encoder layers of ViT to retain the global features, especially the position information, of visible patches. Inspired by He et al. \cite{he2016deep,he2021masked}, we add the embeddings of all image patches before the first encoder layer to the input of the last encoder layer as in Figure~\ref{figure_2}. The input of the last encoder layer $\hat{z}_{L-1}$ is:
\begin{gather}
    \hat{z}_{L-1}=[ \hat{x}_{\scriptscriptstyle PATCH}^{L-1};x_{\scriptscriptstyle DET}^{L-1};x_{\scriptscriptstyle CLS}^{L-1}]\\
    \hat{x}_{\scriptscriptstyle PATCH}^{{L-1}^i}=\begin{cases} x_{\scriptscriptstyle PATCH}^{{L-1}^i} +x_{\scriptscriptstyle PATCH}^{0^i} &, \boldsymbol{if}\;Mask_i=1\\
    x_{\scriptscriptstyle PATCH}^{0^i} &,otherwise \nonumber
    \end{cases}
\end{gather}
where $\hat{x}_{\scriptscriptstyle PATCH}^{L-1}$= $\{ x_{\scriptscriptstyle PATCH}^{{L-1}^i} | i=1,\cdots,N\}$, $x_{\scriptscriptstyle DET}^{L-1}$ and $x_{\scriptscriptstyle CLS}^{L-1}$ are the image patch token, detect token and class token of the last layer, respectively.
Compared with the vanilla ViT model, MG-ViT only introduces an additional mask operation and a residual connection, thus enabling the model to focus on the most task-relevant image patches and to inherit parameters from pre-trained ViT without any other retraining. Therefore, MG-ViT can achieve fast domain adaption for FSL on ViT.

\begin{figure*}[ht]
\begin{center}
\includegraphics[width=1.0\linewidth]{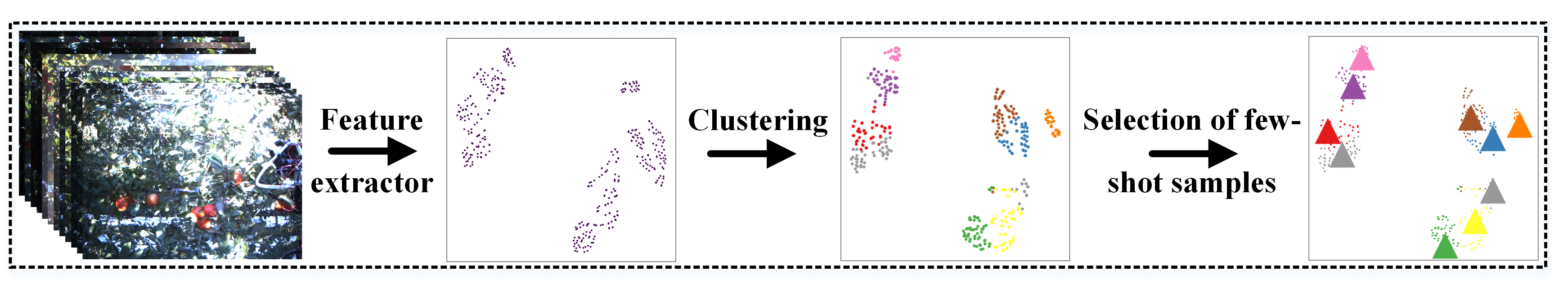}
\caption{A cluster-based active learning method for few-shot sample (triangles in the last sub-figure) selection.}
\label{figure_cluster}
\end{center}
\end{figure*}

\subsection{Generation of Image Patch Mask}
\label{mask}
In order to make better advantage of the prior knowledge and to reduce the deviation between the source domain (base dataset) and target domain (novel dataset), we generate the image patch mask based on the most task-relevant and discriminative patches from the salience map calculated by Grad-CAM \cite{selvaraju2017grad}. Specifically, we first adopt Grad-CAM \cite{selvaraju2017grad} to calculate the salience map of the identified neighborhood samples from the base dataset (detailed in Section~\ref{neighbor}). We then select top \textit{k} salient patches with largest absolute sum values of the gradient of image patch features after patch embedding in the salience map as the most task-relevant and discriminative ones. We finally perform a binary operation on the salience map to label those top \textit{k} salient patches as 1 and the remaining ones as 0. The generation of image patch mask is written as:
\begin{gather}
    g_i = Sum\left|\frac{\partial L(f(x),y)}{\partial x_{\scriptscriptstyle PATCH}^i}\right|,\,i=1,2,\cdots,N
 \label{eq_mask}
 \\
    Mask_i = \begin{cases}
    1&,\boldsymbol{if}\;g_i\;\boldsymbol{in}\;topk\;G\\
    0&,otherwise\\
    \end{cases}
\end{gather}
where $G \in \mathbb{R^N} =\{g_1,g_2,...,g_{N}\}$ is the salience map of patches $x_{\scriptscriptstyle PATCH} = \{ x_{\scriptscriptstyle PATCH} ^1,..., x_{\scriptscriptstyle PATCH} ^{N} \}$, \textit{topk} is the set of selected salient patches and $Mask=\{Mask^1,\cdots,Mask^N\}$ is the binary mask.
Besides the discrete mask generated from the Grad-CAM based salience map, we also generate a continued one based on the center coordinates and designated length and width of the discrete mask in order to provide more contour information of target localization for object detection task (Figure~\ref{figure_2}).

\subsection{Identification of Neighborhood Samples From the Base Dataset}
\label{neighbor}
Identifying neighborhood samples \cite{he2011neighborhood} from the base dataset which are similar to the few-shot samples in the novel dataset for joint fine-tuning can improve the performance for FSL \cite{ge2017borrowing,sbai2020impact,yang2021free}. Inspired by Paul et al. \cite{paul2021deep} using the Gradient Normed and Error $l_2$-Norm scores to identify the important samples which are hard for model training with large loss, we measure the similarity as the negative loss value of the sample in the base dataset, with the model trained on the selected few-shot samples from the novel dataset (detailed in Section~\ref{activelearning}). The similarity of sample $(x_{i},y_{i})\in D_{b}$ with $D_{n}$ is written as:
\begin{gather}
    Sim(x_{i},D_{n}) = -L(f_{\hat{W}}(x_i),y_{i})\\
    \hat{W} = {\mathop{\arg\min}\limits_{W}{L(f_W(x),y)}},\, \forall (x,y)\in D_{n}
\end{gather}
where $L(*)$ is the loss function, $f(*)$ represents the model’s output of input $x$, $W$ is the set of parameters of the model, and $\hat{W}$ is the set of parameters of the model with least loss.

In this way, we effectively combine both the model characteristics and labeling information of the data without performing complex similarity calculations between the two datasets. The proposed method of loss-based image similarity is similar to the anomaly score in anomaly detection \cite{chalapathy2019deep} which measures the distance of data point from the center of a sphere. The lower loss allows the model to focus more on learning the representation of the novel dataset which contributes to the performance for FSL.

\subsection{Active Learning based Few-Shot Sample Selection}\label{activelearning}

To further improve the generalizability of MG-ViT based FSL, we introduce a cluster-based sample selection method to optimally select representative few-shot samples. Considering that the elaborate active learning methods may lead to over-fitting problems, we design a simple and effective method inspired by \cite{cai2011heterogeneous,ge2017borrowing} as illustrated in Figure Figure~\ref{figure_cluster}. We first use a CNN-based model as a feature extractor to extract image features, then perform an unsupervised clustering to identify $k$ clusters as $k$-shot, and finally select the image with the highest node degree in each cluster as the few-shot sample. In this study, we use the pre-trained ResNet-101 \cite{he2016deep} as the backbone to extract features, $k$-means to cluster image features and Euclidean distance as the weight of adjacency.

\subsection{Overall Training Scheme}
\label{training}

\begin{algorithm}[tb]
\caption{Fine-tuning scheme with MG-ViT.}
\label{algorithm_1}
\begin{algorithmic}
    \STATE {\bfseries Input:}  base dataset $D_b$, novel dataset $D_n$ and pretrained model.
    \STATE {\bfseries Initialize:}  model retraining on $D_b$
    \FOR{epoch {\bfseries in} initial fine-tuning step}
        \STATE $train\_one\_epoch(model, (D_b, Mask=None))$
    \ENDFOR
    \WHILE {training}
        \FOR{$(x,y)$ {\bfseries in} $D_b$}
            \STATE $Sim_{x_i} = -loss(f(x_i),y_i)$
        \ENDFOR
    \STATE $D_b^{sub} = topk (Sim_{x_i})$
    \STATE $Mask_b = Grad-CAM(model,D_b^{sub})$ as Eq~\ref{eq_mask}
    \STATE $train\_one\_epoch(model,\! {(\!D_n,\!None\!)\!\cup\!{(\!D_b^{sub},\! Mask_b\!))}}$
    \ENDWHILE
\end{algorithmic}
\end{algorithm}
The overall training scheme of MG-ViT consists of two stages. The first stage is to train on the base dataset only with the vanilla ViT and the second stage is to fine-tune the model on both the base and novel datasets with the proposed MG-ViT. We demonstrate the fine-tuning scheme at the second stage in Algorithm~\ref{algorithm_1}. Note that before we fine-tune MG-ViT with the novel dataset, we first initialize fine-tuning of the model on the base dataset for a few epochs since the change of computational flow of the model at the fine-tuning stage may lead to deviations in the input of the last encoder part.

\section{Experiments}

We evaluate our MG-ViT based on ViT-S  \cite{dosovitskiy2020image} and conduct experiments on both Agri-ImageNet image classification and Apple object detection tasks. We compare our methods with the general fine-tuning based FSL methods \cite{wang2020frustratingly}. We report the average precision (AP) score for object detection task and the average accuracy (ACC) for image classification task on the test split of the novel dataset. 

\subsection{Image Classification in Agri-ImageNet}
\paragraph{Agri-ImageNet}
We carry out the image classification task based on a new Agri-ImageNet dataset. The Agri-ImageNet dataset contains three parent classes including fruit, weed and vegetable. We randomly select the fruit (with 9 sub-classes) and weed (with 8 sub-classes) parent classes as the base dataset and the vegetable (with 4 sub-classes) one as the novel dataset. We perform 4-way few-shot image classification task since there are 4 sub-classes in the vegetable parent class. The base dataset is randomly split into training/test with 75\%/25\%. The remaining data in the novel dataset except for the actively selected few-shot samples is the test split of novel dataset. For the training dataset, Rand-Augment \cite{cubuk2020randaugment}, Random Erasing \cite{zhong2020random}, and RandomResizeCrop to 224 × 224 are applied for data augmentation. For the test dataset, images are only resized and center cropped to 224 × 224. 

\paragraph{Setting}
The ImageNet-1k pre-trained model is firstly trained on the base dataset with the vanilla ViT. We adopt an AdamW optimizer with 200 epochs using a cosine decay learning rate scheduler and 10 epochs of linear warm-up. Batch size is set to 64, initial learning rate is set to 0.0001, and weight decay is set to 0.0001. Then, we fine-tune the model on the few-shot samples in the novel dataset together with the neighborhood samples from the base dataset with MG-ViT. We keep the same setting of regular training except for the initial learning rate to 0.001 and epochs to 30. The cross-entropy loss is adopted and the label smoothing is set to 0.1. The \textit{topk} is set to 7×7 to select the salient patches for mask generation.

\begin{wraptable}{r}{7.5cm}
    \caption{The averaged ACC for different number of shots in image classification task based on Agri-ImageNet.}
    \begin{tabular}{ccccc}
    \toprule
    Backbone  & Method     & 1-shot         & 5-shot        & 10-shot                    \\
    \midrule
    ViT-S     & Ft-full    & 74.9           &90.2           &94.6              \\
    ViT-S     & Ft-part    &60.1            &90.1           &92.6            \\
    MGViT-S   & Ft-part    &65.9            &95.0           & 95.5            \\
    MGViT-S   & Ft-full    &  \textbf{85.6} &\textbf{98.1}  & \textbf{98.5}\\
    \bottomrule
    \end{tabular}
\label{table_1}
\end{wraptable}

\paragraph{Result}
We compare MG-ViT with fine-tuning \cite{wang2020frustratingly} based methods on ViT-S. 
As reported in Table~\ref{table_1}, we see that fine-tuning the whole ViT (Ft-full) instead of only the MLP part (Ft-part) obtains better classification performance. Compared to the fine-tuning based methods (ViT-S), MG-ViT improves 3.9\% from 94.6\% to 98.5\% in 10-shot, 7.9\% from 90.2\% to 98.1\% in 5-shot, and 10.7\% from 74.9\% to 85.6\% in 1-shot, demonstrating the superiority of the proposed MG-ViT in FSL. The patch salience maps are further visualized in Figure~\ref{figure_4} to illustrate that MG-ViT can locate task-relevant patches more accurately compared to the fine-tuning based ViT..

\subsection{Object Detection in ACFR Dataset}

\begin{figure*}[htb]
\begin{center}
\includegraphics[width=0.9\linewidth]{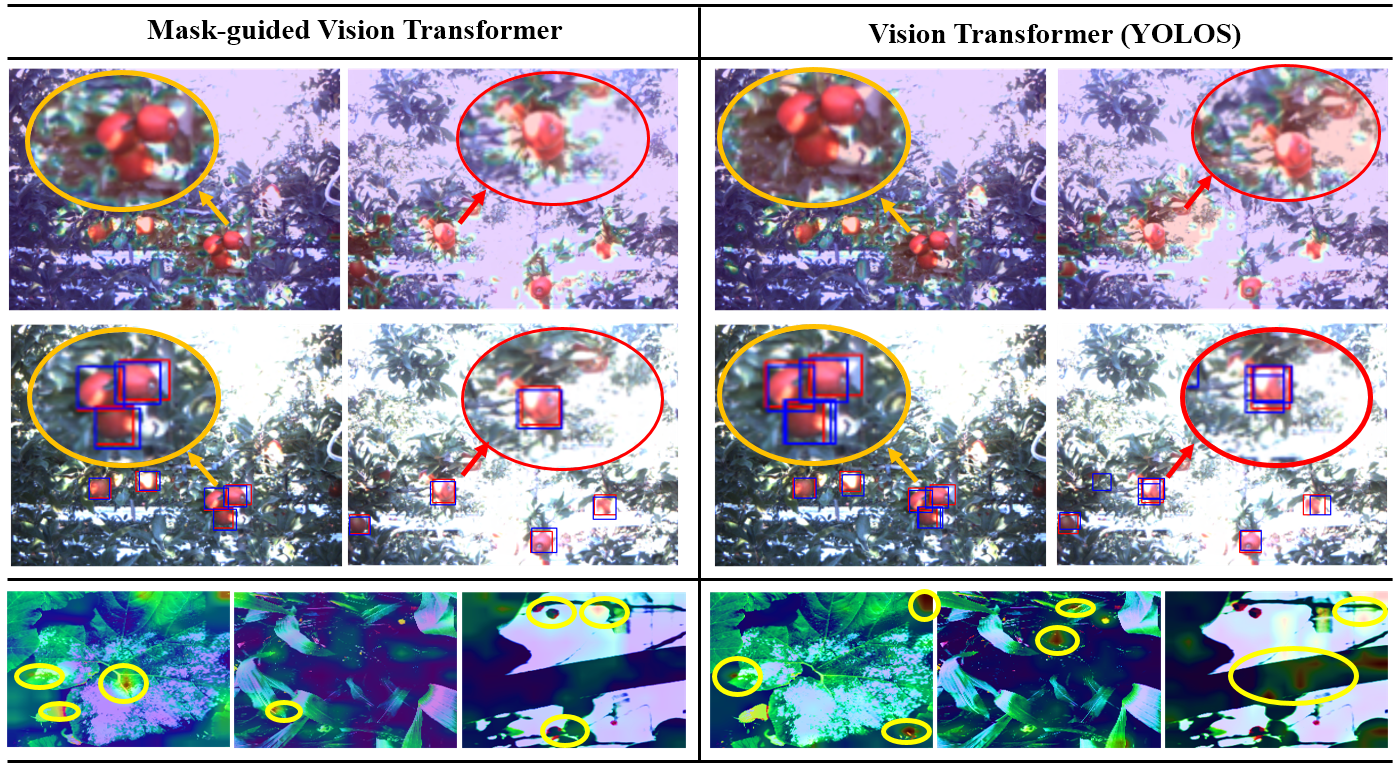}
\caption{\textbf{Top}: Two representative patch salience maps (top) and predict bbox (bottom, {\color{red} red}: ground truth, {\color{blue} blue}: predicted) in MG-ViT and YOLOS in object detection task, respectively. In each sub-figure, \textcolor[RGB]{255,192,0}{yellow} circle highlights the enlarged view of one example salient region. \textbf{Bottom}: Three representative patch salience maps generated by the proposed MG-ViT and ViT in image classification task. The most salient regions are highlighted by \textcolor{yellow}{yellow} circles.}
\label{figure_4}
\end{center}
\vskip -0.15in
\end{figure*}

\paragraph{ACFR Apple}
We apply the object detection task on the apple class of ACFR Orchard Fruit Dataset \cite{bargoti2017image}. The dataset collected in 2016 and 2017 is used as the base and novel dataset, respectively. The base dataset is randomly split into training/validation/test with 64\%/16\%/20\%. The remaining data in the novel dataset except for the few-shot samples is the test split of novel dataset. For the training dataset, data augmentation includes resizing and cropping the input images so that the shortest side is between 432 and 720 pixels and the longest one is at most 960 pixels. For the test dataset, the images are only resized to 720 × 960 and normalized.

\paragraph{Setting}
We follow Fang et al. \cite{fang2021you} and use the provided ImageNet-1k pre-trained ViT-S as the backbone. We continue pre-training the model on the Agri-ImageNet and keep the same setting as the training on the base dataset for better performance. Since the pre-trained model is trained on images with a resolution of 224 × 224 while the ones in the apple dataset are with higher resolution, we adopt a bicubic interpolation \cite{touvron2021training} for position embedding to fit the apple images. AdamW optimizer with a cosine decay learning rate scheduler is employed during the training on the base dataset. Batch size, epochs, initial rate, weight decay are set to 1,200, 0.0001 and 0.0001, respectively. We use the same hyperparameters except for initial rate and epochs set to 0.00001 and 400, respectively during the fine-tuning on the novel dataset. We use the same loss function as in Zhu et al. \cite{zhu2020deformable}. The \textit{topk} is set to 24×32 to select the salient patches for mask generation.

\begin{wraptable}{r}{7.5cm}
    \caption{The AP scores for different number of shots in object detection task based on ACFR Apple dataset.}
    \begin{tabular}{ccccc}
    \toprule
    Backbone            & Method   & 5-shot            & 10-shot           & 30-shot       \\
    \midrule
    ViT-S               & Ft-full  & 47.9              & 56.2              & 74.9          \\
    ViT-S               & Ft-part  & 26.4              & 28.7              & 33.2          \\
    MG-ViT-S            & Ft-full  & \textbf{50.6}     & \textbf{65.3}     & \textbf{76.0} \\
    \bottomrule
    \end{tabular}
\label{table_2}
\end{wraptable}

\paragraph{Result}
We compare the proposed MG-ViT with the baseline method, a pure transformer structural object model of YOLOS-S \cite{fang2021you} with fine-tuning \cite{wang2020frustratingly}. As reported in Table~\ref{table_2}, we see that similarly in Table~\ref{table_1}, fine-tuning the full network (Ft-full) performs better than only fine-tuning the backbone (Ft-part) of the model, which is contrary to the conclusion of CNN-based few-shot object detection. By performing the mask operation, our MG-ViT outperforms the general fine-tuning based method (ViT-S) by 2.7\% from 47.9\% to 50.6\% in 5-shot, 9.1\% from 56.2\% to 65.3\% in 10-shot, and 1.1\% from 74.9\% to 76.0\% in 30-shot. The patch salience maps of MG-ViT and YOLOS are further visualized and compared in Figure~\ref{figure_4} to illustrate that MG-ViT can locate the object (i.e., apple) more accurately compared to YOLOS.

\section{Ablation Study}

\begin{table*}[t]
\caption{The averaged ACC in image classification task and AP in object detection task for different number of shots based on different combinations of model design.}
\label{table_3}
\begin{tabular}{ccccccc}
\toprule
Task & Active Learning & Neighborhood  & Masked    & 1-shot    & 5-shot    & 10-shot \\
\midrule
\multirow{4}{*}{Clasiffication}   
&           & $\surd$       & $\surd$   & 84.6              &96.5              & 96.6    \\
&  $\surd$  &               &           & 77.0              &90.4              & 96.7    \\
&  $\surd$  & $\surd$       &           & 82.1              &97.8              &97.9  \\
&  $\surd$  & $\surd$       & $\surd$   &  {\textbf{85.6}}  &{\textbf{98.1}}   &{\textbf{98.5}}\\
\midrule
Task& Active Learning & Neighborhood  & Masked    & 5-shot     & 10-shot    & 30-shot \\
\midrule
\multirow{4}{*}{Detection}
&                & $\surd$       & $\surd$   & 47.2          & 56.3          & 75.1      \\
&$\surd$         &               &           & 48.2          & 63.2          & 72.9      \\
&$\surd$         & $\surd$       &           & 48.6          & 64.1          & 75.3      \\
&$\surd$         & $\surd$       & $\surd$   & \textbf{50.6} & \textbf{65.3} & \textbf{76.0} \\
\bottomrule
\end{tabular}
\end{table*}

\subsection{Effect of Active Learning based Few-Shot Sample Selection}
We compare the effect of active learning based few-shot sample selection with randomly selected few-shot samples in MG-ViT. As reported in Figure~\ref{table_3}, for image classification task, active learning based few-shot sample selection improves the accuracy by 1.0\% from 84.6\% to 85.6\% in 1-shot, 1.6\% from 96.5\% to 98.1\% in 5-shot, and 1.9\% from 96.6\% to 98.5\% in 10-shot. Similarly, for object detection task, it improves the accuracy by 3.4\% from 47.2\% to 50.6\% in 5-shot, 9.0\% from 56.3\% to 65.3\% in 10-shot, and 0.9\% from 75.1\% to 76.0\% in 30-shot.

\subsection{Effect of Neighborhood Samples Identification}
We compare the effect of adopting identified neighborhood samples with randomly selected images from the base dataset in MG-ViT. As reported in Figure~\ref{table_3}, for image classification task, adopting identified neighborhood samples for joint fine-tuning with the novel dataset improves the accuracy by 5.1\% from 77.0\% to 82.1\% in 1-shot, 7.4\% from 90.4\% to 97.8\% in 5-shot, and 1.2\% from 96.7\% to 97.9\% in 10-shot. Similarly, for object detection task, it improves the accuracy by 0.4\% from 48.2\% to 48.6\% in 5-shot, 0.9\% from 63.2\% to 64.1\% in 10-shot, and 2.4\% from 72.9\% to 75.3\% in 30-shot.


\begin{wraptable}{r}{7.5cm}
    \caption{The performances of discrete and continued masks on 5-shot image classification and 10-shot object detection with MG-ViT.}
    \begin{tabular}{ccc}
    \toprule
    Mask Shape     & Discrete  & Continued               \\
    \midrule
    5-shot image classification   & \textbf{98.7}           & 98.1                   \\
    10-shot object detection&      63.1           & \textbf{65.3}                \\
    \bottomrule
    \end{tabular}
\end{wraptable}

\subsection{Effect of Discrete or Continued Mask}
We evaluate the effect of using a discrete or continued mask in MG-ViT. As reported in Table 4, the discrete mask performs better for image classification while the continued one performs better for object detection. This difference may derive from the fact that the continued mask could provide more contour information of target localization which benefits for object detection, while the discrete mask may provide more global semantic information for image classification task. We leave exploration of different types of masks for different downstream tasks in FSL to future work.

\section{Conclusion}

We propose a novel mask-guided vision transformer (MG-ViT) to guide ViT more effectively learn from the task-relevant prior knowledge for few-shot learning. By simply adding an image patch mask operation and a residual connection to the vanilla ViT, MG-ViT significantly outperforms the general fine-tuning based methods for FSL. Our results are in agreement with ViT that \emph{learning the relevant patterns directly from data is sufficient, even beneficial}. To further improve FSL, we also introduce an effective active learning based few-shot sample selection method. Our two-stage fine-tuning based framework could be widely applied on different downstream tasks such as image classification and object detection in this study. In general, MG-ViT provides a concrete approach towards generalizing data-intensive and large-scale deep learning models for FSL. In the future, we can systematically explore the effectiveness of different types (e.g., size, shape, data modality, etc.) of image patch masks for different downstream tasks. Further combing the proposed mask operation with other models and modalities in other tasks such as NLP is another exciting direction of future work.


\begin{thebibliography}{10}

\bibitem{wang2020generalizing}
Yaqing Wang, Quanming Yao, James~T Kwok, and Lionel~M Ni.
\newblock Generalizing from a few examples: A survey on few-shot learning.
\newblock {\em ACM Computing Surveys (CSUR)}, 53(3):1--34, 2020.

\bibitem{yang2021free}
Shuo Yang, Lu~Liu, and Min Xu.
\newblock Free lunch for few-shot learning: Distribution calibration.
\newblock {\em arXiv preprint arXiv:2101.06395}, 2021.

\bibitem{yan2022budget}
Shipeng Yan, Songyang Zhang, and Xuming He.
\newblock Budget-aware few-shot learning via graph convolutional network.
\newblock {\em arXiv preprint arXiv:2201.02304}, 2022.

\bibitem{ma2021few}
Yuqing Ma, Wei Liu, Shihao Bai, Qingyu Zhang, Aishan Liu, Weimin Chen, and
  Xianglong Liu.
\newblock Few-shot visual learning with contextual memory and fine-grained
  calibration.
\newblock In {\em Proceedings of the Twenty-Ninth International Conference on
  International Joint Conferences on Artificial Intelligence}, pages 811--817,
  2021.

\bibitem{zhang2021rethinking}
Hongguang Zhang, Piotr Koniusz, Songlei Jian, Hongdong Li, and Philip~HS Torr.
\newblock Rethinking class relations: Absolute-relative supervised and
  unsupervised few-shot learning.
\newblock In {\em Proceedings of the IEEE/CVF Conference on Computer Vision and
  Pattern Recognition}, pages 9432--9441, 2021.

\bibitem{nakamura2019revisiting}
Akihiro Nakamura and Tatsuya Harada.
\newblock Revisiting fine-tuning for few-shot learning.
\newblock {\em arXiv preprint arXiv:1910.00216}, 2019.

\bibitem{wang2020frustratingly}
Xin Wang, Thomas~E Huang, Trevor Darrell, Joseph~E Gonzalez, and Fisher Yu.
\newblock Frustratingly simple few-shot object detection.
\newblock {\em arXiv preprint arXiv:2003.06957}, 2020.

\bibitem{cai2020cross}
John Cai and Sheng~Mei Shen.
\newblock Cross-domain few-shot learning with meta fine-tuning.
\newblock {\em arXiv preprint arXiv:2005.10544}, 2020.

\bibitem{dosovitskiy2020image}
Alexey Dosovitskiy, Lucas Beyer, Alexander Kolesnikov, Dirk Weissenborn,
  Xiaohua Zhai, Thomas Unterthiner, Mostafa Dehghani, Matthias Minderer, Georg
  Heigold, Sylvain Gelly, et~al.
\newblock An image is worth 16x16 words: Transformers for image recognition at
  scale.
\newblock {\em arXiv preprint arXiv:2010.11929}, 2020.

\bibitem{khan2021transformers}
Salman Khan, Muzammal Naseer, Munawar Hayat, Syed~Waqas Zamir, Fahad~Shahbaz
  Khan, and Mubarak Shah.
\newblock Transformers in vision: A survey.
\newblock {\em arXiv preprint arXiv:2101.01169}, 2021.

\bibitem{luo2021rectifying}
Xu~Luo, Longhui Wei, Liangjian Wen, Jinrong Yang, Lingxi Xie, Zenglin Xu, and
  Qi~Tian.
\newblock Rectifying the shortcut learning of background for few-shot learning.
\newblock {\em Advances in Neural Information Processing Systems}, 34, 2021.

\bibitem{selvaraju2017grad}
Ramprasaath~R Selvaraju, Michael Cogswell, Abhishek Das, Ramakrishna Vedantam,
  Devi Parikh, and Dhruv Batra.
\newblock Grad-cam: Visual explanations from deep networks via gradient-based
  localization.
\newblock In {\em Proceedings of the IEEE international conference on computer
  vision}, pages 618--626, 2017.

\bibitem{cohn1994improving}
David Cohn, Les Atlas, and Richard Ladner.
\newblock Improving generalization with active learning.
\newblock {\em Machine learning}, 15(2):201--221, 1994.

\bibitem{chitta2021training}
Kashyap Chitta, Jos{\'e}~M {\'A}lvarez, Elmar Haussmann, and Cl{\'e}ment
  Farabet.
\newblock Training data subset search with ensemble active learning.
\newblock {\em IEEE Transactions on Intelligent Transportation Systems}, 2021.

\bibitem{konyushkova2017learning}
Ksenia Konyushkova, Raphael Sznitman, and Pascal Fua.
\newblock Learning active learning from data.
\newblock {\em arXiv preprint arXiv:1703.03365}, 2017.

\bibitem{vaswani2017attention}
Ashish Vaswani, Noam Shazeer, Niki Parmar, Jakob Uszkoreit, Llion Jones,
  Aidan~N Gomez, {\L}ukasz Kaiser, and Illia Polosukhin.
\newblock Attention is all you need.
\newblock In {\em Advances in neural information processing systems}, pages
  5998--6008, 2017.

\bibitem{touvron2021training}
Hugo Touvron, Matthieu Cord, Matthijs Douze, Francisco Massa, Alexandre
  Sablayrolles, and Herv{\'e} J{\'e}gou.
\newblock Training data-efficient image transformers \& distillation through
  attention.
\newblock In {\em International Conference on Machine Learning}, pages
  10347--10357. PMLR, 2021.

\bibitem{yuan2021incorporating}
Kun Yuan, Shaopeng Guo, Ziwei Liu, Aojun Zhou, Fengwei Yu, and Wei Wu.
\newblock Incorporating convolution designs into visual transformers.
\newblock {\em arXiv preprint arXiv:2103.11816}, 2021.

\bibitem{li2021localvit}
Yawei Li, Kai Zhang, Jiezhang Cao, Radu Timofte, and Luc Van~Gool.
\newblock Localvit: Bringing locality to vision transformers.
\newblock {\em arXiv preprint arXiv:2104.05707}, 2021.

\bibitem{zhang2021aggregating}
Zizhao Zhang, Han Zhang, Long Zhao, Ting Chen, and Tomas Pfister.
\newblock Aggregating nested transformers.
\newblock {\em arXiv preprint arXiv:2105.12723}, 2021.

\bibitem{chen2021vision}
Xiangning Chen, Cho-Jui Hsieh, and Boqing Gong.
\newblock When vision transformers outperform resnets without pretraining or
  strong data augmentations.
\newblock {\em arXiv preprint arXiv:2106.01548}, 2021.

\bibitem{hassani2021escaping}
Ali Hassani, Steven Walton, Nikhil Shah, Abulikemu Abuduweili, Jiachen Li, and
  Humphrey Shi.
\newblock Escaping the big data paradigm with compact transformers.
\newblock {\em arXiv preprint arXiv:2104.05704}, 2021.

\bibitem{he2021masked}
Kaiming He, Xinlei Chen, Saining Xie, Yanghao Li, Piotr Doll{\'a}r, and Ross
  Girshick.
\newblock Masked autoencoders are scalable vision learners.
\newblock {\em arXiv preprint arXiv:2111.06377}, 2021.

\bibitem{vanschoren2018meta}
Joaquin Vanschoren.
\newblock Meta-learning: A survey.
\newblock {\em arXiv preprint arXiv:1810.03548}, 2018.

\bibitem{liu2020universal}
Lu~Liu, William Hamilton, Guodong Long, Jing Jiang, and Hugo Larochelle.
\newblock A universal representation transformer layer for few-shot image
  classification.
\newblock {\em arXiv preprint arXiv:2006.11702}, 2020.

\bibitem{gan2021transformer}
Tao Gan, Weichao Li, Yuanzhe Lu, and Yanmin He.
\newblock Transformer-based few-shot learning for image classification.
\newblock In {\em International Conference on Artificial Intelligence for
  Communications and Networks}, pages 68--74. Springer, 2021.

\bibitem{chen2021sparse}
Haoxing Chen, Huaxiong Li, Yaohui Li, and Chunlin Chen.
\newblock Sparse spatial transformers for few-shot learning.
\newblock {\em arXiv preprint arXiv:2109.12932}, 2021.

\bibitem{fang2021you}
Yuxin Fang, Bencheng Liao, Xinggang Wang, Jiemin Fang, Jiyang Qi, Rui Wu,
  Jianwei Niu, and Wenyu Liu.
\newblock You only look at one sequence: Rethinking transformer in vision
  through object detection.
\newblock {\em arXiv preprint arXiv:2106.00666}, 2021.

\bibitem{he2016deep}
Kaiming He, Xiangyu Zhang, Shaoqing Ren, and Jian Sun.
\newblock Deep residual learning for image recognition.
\newblock In {\em Proceedings of the IEEE conference on computer vision and
  pattern recognition}, pages 770--778, 2016.

\bibitem{he2011neighborhood}
Qiang He, Zongxia Xie, Qinghua Hu, and Congxin Wu.
\newblock Neighborhood based sample and feature selection for svm
  classification learning.
\newblock {\em Neurocomputing}, 74(10):1585--1594, 2011.

\bibitem{ge2017borrowing}
Weifeng Ge and Yizhou Yu.
\newblock Borrowing treasures from the wealthy: Deep transfer learning through
  selective joint fine-tuning.
\newblock In {\em Proceedings of the IEEE conference on computer vision and
  pattern recognition}, pages 1086--1095, 2017.

\bibitem{sbai2020impact}
Othman Sbai, Camille Couprie, and Mathieu Aubry.
\newblock Impact of base dataset design on few-shot image classification.
\newblock In {\em European Conference on Computer Vision}, pages 597--613.
  Springer, 2020.

\bibitem{paul2021deep}
Mansheej Paul, Surya Ganguli, and Gintare~Karolina Dziugaite.
\newblock Deep learning on a data diet: Finding important examples early in
  training.
\newblock {\em Advances in Neural Information Processing Systems}, 34, 2021.

\bibitem{chalapathy2019deep}
Raghavendra Chalapathy and Sanjay Chawla.
\newblock Deep learning for anomaly detection: A survey.
\newblock {\em arXiv preprint arXiv:1901.03407}, 2019.

\bibitem{cai2011heterogeneous}
Xiao Cai, Feiping Nie, Heng Huang, and Farhad Kamangar.
\newblock Heterogeneous image feature integration via multi-modal spectral
  clustering.
\newblock In {\em CVPR 2011}, pages 1977--1984. IEEE, 2011.

\bibitem{cubuk2020randaugment}
Ekin~D Cubuk, Barret Zoph, Jonathon Shlens, and Quoc~V Le.
\newblock Randaugment: Practical automated data augmentation with a reduced
  search space.
\newblock In {\em Proceedings of the IEEE/CVF Conference on Computer Vision and
  Pattern Recognition Workshops}, pages 702--703, 2020.

\bibitem{zhong2020random}
Zhun Zhong, Liang Zheng, Guoliang Kang, Shaozi Li, and Yi~Yang.
\newblock Random erasing data augmentation.
\newblock In {\em Proceedings of the AAAI Conference on Artificial
  Intelligence}, volume~34, pages 13001--13008, 2020.

\bibitem{bargoti2017image}
Suchet Bargoti and James~P Underwood.
\newblock Image segmentation for fruit detection and yield estimation in apple
  orchards.
\newblock {\em Journal of Field Robotics}, 34(6):1039--1060, 2017.

\bibitem{zhu2020deformable}
Xizhou Zhu, Weijie Su, Lewei Lu, Bin Li, Xiaogang Wang, and Jifeng Dai.
\newblock Deformable detr: Deformable transformers for end-to-end object
  detection.
\newblock {\em arXiv preprint arXiv:2010.04159}, 2020.

\end{thebibliography}
\end{document}